\documentclass{article} 
\usepackage{iclr2027_conference,times}

\usepackage{amsmath,amsfonts,bm}

\def\eqref#1{equation~\ref{#1}}
\def\Eqref#1{Equation~\ref{#1}}

\def\1{\bm{1}}

\DeclareMathAlphabet{\mathsfit}{\encodingdefault}{\sfdefault}{m}{sl}
\SetMathAlphabet{\mathsfit}{bold}{\encodingdefault}{\sfdefault}{bx}{n}

\DeclareMathOperator*{\argmax}{arg\,max}

\usepackage{graphicx,wrapfig}
\usepackage{booktabs}
\usepackage{array}
\usepackage{multirow}
\usepackage{tabularx}
\usepackage[table]{xcolor}
\usepackage{placeins}
\usepackage{needspace}
\usepackage{algorithm}
\usepackage{algpseudocode}
\usepackage[hidelinks]{hyperref}
\usepackage{url}
\usepackage{fontawesome5}

\definecolor{studentfour}{rgb}{0.26741028,0.46926880,0.56454468}
\definecolor{studenteight}{rgb}{0.57954407,0.31268311,0.54252625}
\definecolor{teachergroup}{gray}{0.92}
\definecolor{aopdgain}{HTML}{276449}
\newcommand{\qwenmark}{\raisebox{-0.12em}{\includegraphics[width=0.95em,height=0.85em,keepaspectratio]{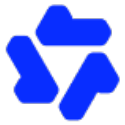}}\hspace{0.25em}}
\newcommand{\deepseekmark}{\raisebox{-0.12em}{\includegraphics[width=0.95em,height=0.85em,keepaspectratio]{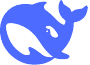}}\hspace{0.25em}}
\newcommand{\scoresep}{\kern0.3pt/\kern0.3pt}
\newcommand{\avgcell}[2]{\makebox[4.3em][c]{#1\if\relax\detokenize{#2}\relax\else\rlap{\,\textsuperscript{\textcolor{aopdgain}{\scriptsize +#2}}}\fi}}

\newcommand{\projectpageurl}{https://thu-nics.github.io/ActionOPD/}
\newcommand{\codeurl}{https://github.com/thu-nics/ActionOPD}
\newcommand{\modelsurl}{https://huggingface.co/collections/nics-efc/actionopd}
\newcommand{\iconcenter}[1]{\raisebox{\dimexpr0.5\depth-0.5\height\relax}{#1}}
\newcommand{\resourcelink}[3]{%
    \href{#1}{%
        {\small\sffamily\textcolor{studentfour!40!black}{%
            \mbox{\raisebox{0.5ex}{\makebox[1.2em][c]{\iconcenter{#2}}}\hspace{0.35em}\raisebox{-1.5pt}[\height][\depth]{#3}}}}%
    }%
}
\newcommand{\hflogo}{\includegraphics[height=11.5pt]{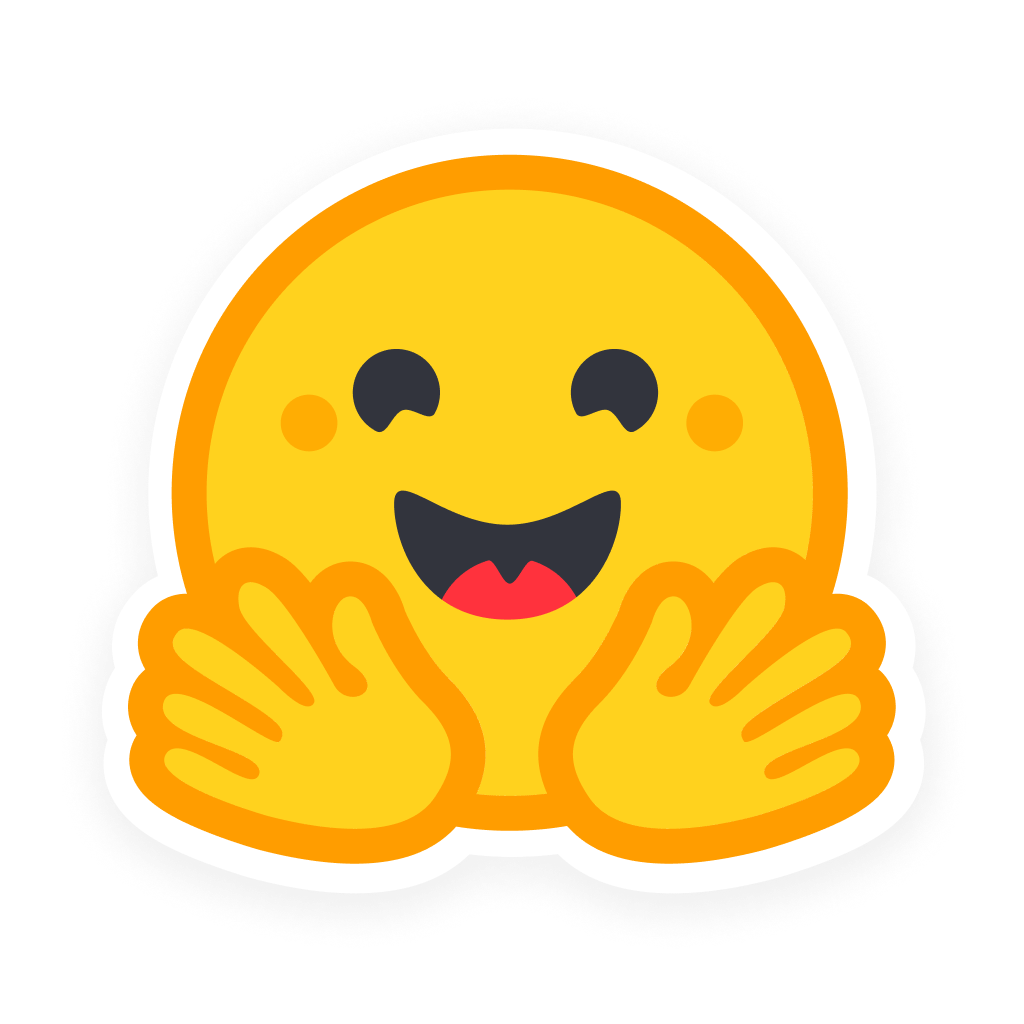}}

\hypersetup{pdftitle={ActionOPD: Distilling Decisions, Not Tokens}}
\title{ActionOPD: Distilling Decisions, Not Tokens}

\author{%
\begin{minipage}[c]{\dimexpr\textwidth-2\tabcolsep\relax}
    \normalfont
    \textbf{\href{mailto:yuanhuining0@gmail.com}{Huining Yuan}\textsuperscript{*}, Xiangmin Yi\textsuperscript{*}, Zelai Xu, Zixu Wang,}\\
    \textbf{Xinlei Chen, \href{mailto:xiaoping.zhang@sz.tsinghua.edu.cn}{Xiao-Ping Zhang}\textsuperscript{$\dagger$}, \href{mailto:yu-wang@tsinghua.edu.cn}{Yu Wang}\textsuperscript{$\dagger$}, \href{mailto:yuchao@sz.tsinghua.edu.cn}{Chao Yu}\textsuperscript{$\dagger$}}\\
    \makebox[\linewidth][l]{Tsinghua University}\\
    \makebox[\linewidth][l]{%
        {\small \textsuperscript{*}Equal contribution.\quad \textsuperscript{$\dagger$}Corresponding authors.}}\\
    \makebox[\linewidth][l]{%
        \resourcelink{\projectpageurl}{\faGlobe}{Project Page}\hspace{1em}%
        \resourcelink{\codeurl}{\faGithub}{Code}\hspace{1em}%
        \resourcelink{\modelsurl}{\hflogo}{Models}}
\end{minipage}}

\iclrfinalcopy 
\begin{document}

\maketitle

\begin{abstract}
Distilling capable language-model agents into smaller models requires aligning the student's behavior with the teacher's. This raises a fundamental question: at which states and at what granularity should this alignment be performed? Offline imitation provides dense but token-specific supervision and can suffer from state-distribution mismatch. Token-level on-policy distillation (OPD) addresses this mismatch but typically requires teacher token probabilities and compatible tokenization. We argue that semantic actions provide a natural unit of supervision for agent distillation. We introduce ActionOPD, an action-level on-policy distillation method motivated by maximizing expected teacher-action likelihood at student-encountered states. At each such state, ActionOPD samples actions from both the student and a black-box teacher, rewarding student actions according to their semantic agreement with the sampled teacher actions. Through policy optimization, ActionOPD aligns the student's action-level behavior with the teacher's decisions without requiring reproduction of teacher-written reasoning or access to teacher token probabilities. Training Qwen3-4B and Qwen3-8B with two teachers on diverse, curated tool-use tasks yields average gains of 3.3--6.9 percentage points across four out-of-distribution benchmarks. ActionOPD scales favorably with student size and teacher capability, without requiring the teacher and student to share a model family or tokenizer. On $\tau^2$-bench, ActionOPD achieves competitive performance against offline distillation and token-level OPD. We further explore Self-ActionOPD with a same-model teacher and privileged context. These results establish ActionOPD as a promising approach to agent distillation.
\end{abstract}

\section{Introduction}

\begin{wrapfigure}[19]{r}{0.4\textwidth}
    \vspace{-1.35\baselineskip}
    \centering
    \includegraphics[width=\linewidth]{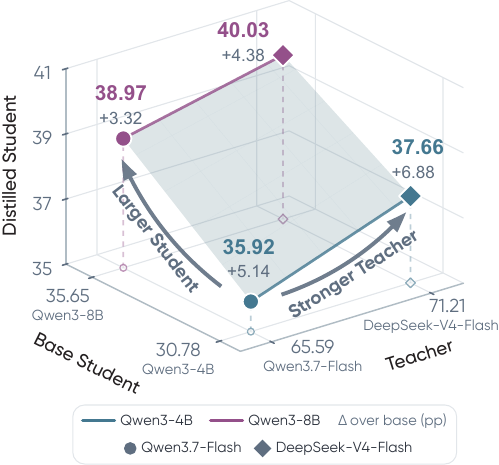}
    \caption{Student--teacher scaling of ActionOPD. Larger students attain higher final averages. DeepSeek-V4-Flash yields higher student averages than Qwen3.7-Flash.}
    \label{fig:teaser}
\end{wrapfigure}

Language models are increasingly deployed as agents that communicate with users, maintain state, and act through external tools~\citep{yao2023react,yao2025taubench}. In these interactions, an agent's decisions shape subsequent observations and the states it encounters. Distilling capable but expensive frontier agents into smaller models therefore raises two questions: \emph{at which states} should the teacher supervise the student, and \emph{at what granularity} should their behavior be aligned?

\begin{figure}[t]
    \centering
    \includegraphics[width=\linewidth]{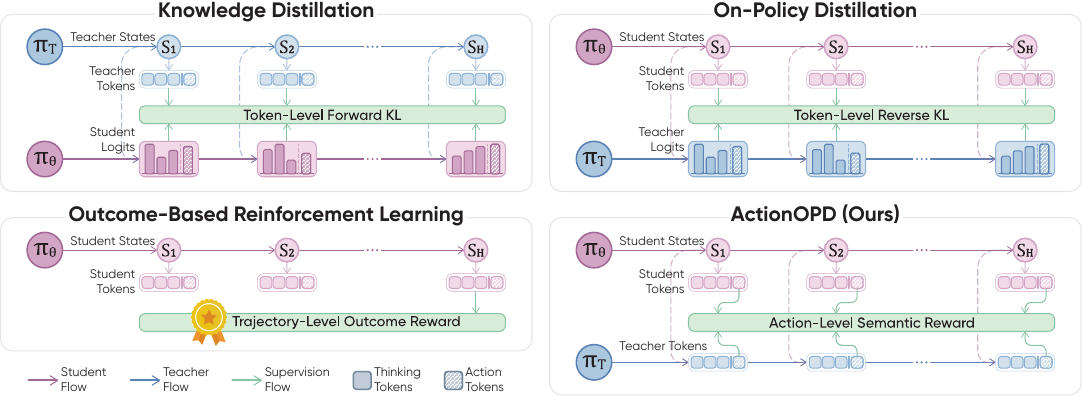}
    \caption{Supervision for agent learning. Offline distillation (KD) imitates teacher tokens at teacher-induced states. Token-level OPD compares token distributions along student-generated interactions. Outcome-based RL uses terminal feedback. ActionOPD instead rewards semantic agreement with teacher actions at student-encountered states. Here, we sample only one response per state for demonstration. For implementation, we expand the sampled teacher multiset and student state group.}
    \label{fig:main-comparison}
\end{figure}

Distillation methods address these questions differently. Offline distillation provides dense supervision by fitting teacher-generated sequences~\citep{kim2016sequence}, but can face teacher--student state-distribution mismatch~\citep{ross2011reduction}. Token-level on-policy distillation (OPD) supervises student-generated sequences using teacher token probabilities, reducing mismatch while typically requiring compatible tokenization~\citep{agarwal2024onpolicy,gu2024minillm}. Both supervise tokens without explicitly matching action semantics (Figure~\ref{fig:main-comparison}).

Our key observation is that \emph{semantic actions} provide a natural unit of supervision for agent distillation. A tool call expresses a decision through its tool name and arguments; a natural-language response conveys communicative intent and information. The same decision can be realized through different wording and reasoning. This shifts the objective from reproducing a particular teacher sequence to increasing the likelihood of its underlying action.

We introduce \textbf{ActionOPD}, an action-level on-policy distillation method motivated by maximizing expected teacher-action likelihood at student-encountered states. Here, likelihood is the total probability mass of student outputs that semantically match a teacher action, not the probability of reproducing one teacher-written sequence. This objective yields a REINFORCE-form policy gradient with action-level semantic agreement as reward~\citep{williams1992simple}. In practice, ActionOPD samples actions from the teacher and responses from the current student policy, rewarding student responses according to their semantic agreement with teacher actions. It requires only generation access to a black-box teacher, not teacher token probabilities or tokenizer alignment. The resulting action-level rewards train the student's own reasoning and action tokens without imposing teacher-written reasoning as a token-level imitation target.

We train Qwen3-4B and Qwen3-8B students using teachers from two model families on curated tool-use tasks from Agent World Model (AWM) and EnvScaler~\citep{wang2026awm,song2026envscaler}. All four configurations improve average performance across four out-of-distribution agent benchmarks. ActionOPD scales favorably with student size and teacher capability: larger students achieve higher final averages, and DeepSeek-V4-Flash yields higher student averages than Qwen3.7-Flash at both sizes (Figure~\ref{fig:teaser}). Comparisons on $\tau^2$-bench~\citep{barres2025tau2} show competitive performance against offline distillation and token-level OPD. Ablation studies support the effectiveness of semantic action supervision. We further explore Self-ActionOPD, in which a same-model teacher receives privileged task information. This extension suggests a path toward agent self-improvement through action-level self-distillation.

\Needspace{5\baselineskip}
\paragraph{Contributions.}
Our contributions are threefold:
\begin{itemize}
    \item We introduce ActionOPD, an action-level on-policy distillation method that decouples teacher decision supervision from student token realization, enabling distillation from black-box teachers without teacher token probabilities or tokenizer alignment.
    \item We demonstrate ActionOPD's applicability across teacher families and empirical scaling with student size and teacher capability. All tested configurations improve the four-benchmark average over their corresponding base models by 3.3--6.9 percentage points.
    \item We show that ActionOPD achieves competitive performance against offline distillation and token-level OPD while requiring only generation access to the teacher.
\end{itemize}

\section{ActionOPD}
\label{sec:method}

A language-model agent alternates between generating responses and receiving environment observations. At step $t$, its visible state $s_t$ contains the task, tools, and retained interaction history. Its response $y_t$ realizes a semantic action $a_t=\phi_{s_t}(y_t)$. Here, \emph{action level} refers to the semantics of a tool call or user-facing message, rather than the token sequence used to express it. The function $\phi_s$ maps responses to semantic actions at state $s$. We use $\pi_\theta$ for student probabilities over semantic actions, complete responses, and tokens, distinguished by their arguments, and $\pi_{\mathrm T}$ for the teacher. Student and teacher receive the same public context and tools through their native prompts.

To align the student's action-level behavior with the teacher, we maximize expected teacher-action likelihood at student-encountered states:
\begin{equation}
    \max_\theta\;
    \mathbb E_{s\sim d_{\mathrm S},\,a^{\mathrm T}\sim\pi_{\mathrm T}(\cdot\mid s)}
    \left[\pi_\theta(a^{\mathrm T}\mid s)\right].
    \label{eq:distillation-objective}
\end{equation}
Denote this expectation by $J_{\mathrm{ActionOPD}}(\theta)$. Here, $d_{\mathrm S}$ denotes the distribution of student-encountered states collected by the rollout procedure in Section~\ref{sec:state-group}, held fixed during each policy update.

\subsection{From action likelihood to policy gradients}
\label{sec:action-likelihood}

To optimize \Eqref{eq:distillation-objective}, we express the likelihood of a teacher action as the total probability mass of student responses that realize it:
\begin{equation}
    \pi_\theta(a^{\mathrm T}\mid s)
    =\sum_y\pi_\theta(y\mid s)\,\mathbf 1\{\phi_s(y)=a^{\mathrm T}\}.
    \label{eq:action-likelihood}
\end{equation}
Here, $\mathbf 1\{\phi_s(y)=a^{\mathrm T}\}$ is the semantic-agreement indicator: it equals one when $y$ realizes the teacher action and zero otherwise. Agreement is defined by action semantics, not textual identity. Substituting \Eqref{eq:action-likelihood} into \Eqref{eq:distillation-objective} and exchanging expectations yields an equivalent reward objective:
\begin{equation}
\begin{aligned}
    r_{\mathrm T}(s,a)
      &=\mathbb E_{a^{\mathrm T}\sim\pi_{\mathrm T}(\cdot\mid s)}
        [\mathbf 1\{a=a^{\mathrm T}\}],\\
    J_{\mathrm{ActionOPD}}(\theta)
      &=\mathbb E_{s\sim d_{\mathrm S},\,y\sim\pi_\theta(\cdot\mid s)}
        [r_{\mathrm T}(s,\phi_s(y))].
\end{aligned}
\label{eq:agreement-objective}
\end{equation}
For a fixed state and student action, $r_{\mathrm T}$ averages binary semantic agreement over the teacher's action distribution. With the state distribution, teacher policy, and semantic mapping held fixed during the update, the score-function identity yields
\begin{equation}
    \nabla_\theta J_{\mathrm{ActionOPD}}(\theta)
    =\mathbb E_{s,a^{\mathrm T},y}
      \!\left[\mathbf 1\{\phi_s(y)=a^{\mathrm T}\}
      \nabla_\theta\log\pi_\theta(y\mid s)\right],
    \label{eq:action-policy-gradient}
\end{equation}
where the variables follow the distributions above. This is precisely the REINFORCE policy-gradient form~\citep{williams1992simple}, with semantic action agreement as reward (Appendix~\ref{app:derivation}).

This gradient can be estimated from sampled teacher actions and student responses using student log-probabilities, without teacher token probabilities or tokenizer alignment. Since $\log\pi_\theta(y\mid s)=\sum_{\ell=1}^{|y|}\log\pi_\theta(y_\ell\mid s,y_{<\ell})$, where $\ell$ indexes response tokens, a single semantic-agreement reward weights every token's contribution to the policy gradient. The supervision is therefore action-level, while optimization covers the student's own reasoning and action tokens without imposing teacher-written reasoning as a hard target.

\subsection{Semantic matching and reward design}
\label{sec:semantic-supervision}

We estimate the expected agreement reward $r_{\mathrm T}(s,\phi_s(y))$ using sampled teacher responses and a practical semantic matcher. At state $s$, we request $K=3$ teacher responses realizing $a_k^{\mathrm T}=\phi_s(y_k^{\mathrm T})$ and retain the multiset $\mathcal M_{\mathrm T}(s)=[y^{\mathrm T}_1,\ldots,y^{\mathrm T}_K]$. The matcher $m_s(y,y_k^{\mathrm T})\in\{0,1\}$ approximates $\mathbf 1\{\phi_s(y)=a_k^{\mathrm T}\}$. Invalid outputs never match. Define
\begin{equation}
    c(s,y)=\sum_{k=1}^{K}m_s(y,y^{\mathrm T}_k).
    \label{eq:teacher-count}
\end{equation}
Repeated teacher samples contribute separately to the count. Under independent teacher draws, $c(s,y)/K$ is an unbiased estimate of the probability that the matcher judges $y$ to agree with a sampled teacher response. With accurate matching, this probability equals $r_{\mathrm T}(s,\phi_s(y))$. Appendix~\ref{app:protocol} details sample retention and caching.

The matcher compares action content, not reasoning (Appendix~\ref{app:matching}). Tool names and canonicalized arguments are checked programmatically, with a frozen LLM assessing remaining argument differences. For example, \texttt{get\_weather} calls with \texttt{city="Paris"} and \texttt{unit="celsius"} are equivalent regardless of argument-key order. For communicative responses, we accept normalized exact matches and otherwise ask the LLM to compare intent and factual content. At the same state, ``Basic-economy tickets cannot be changed'' and ``Changes are not permitted for basic-economy tickets'' convey the same restriction. Each candidate is compared with every teacher sample.

Although $c/K$ estimates agreement, three samples provide only coarse frequency estimates. To limit sensitivity to sampled counts, we assign a unit reward to the first match and a smaller frequency bonus to each additional match. For $K>1$, we use
\begin{equation}
    r_\lambda(s,y)=
    \begin{cases}
        -1, & y\text{ contains an invalid action},\\
        0, & y\text{ contains a valid action and }c(s,y)=0,\\
        1+\lambda\dfrac{c(s,y)-1}{K-1}, & c(s,y)>0.
    \end{cases}
    \label{eq:calibrated-reward}
\end{equation}
With $\lambda=0.5$ and $[A,A,B]$, a response matching $A$ earns $1.25$ and one matching $B$ earns $1$: the additional $A$ match contributes $0.25$, compared with $1$ for the first match. Invalid outputs incur a separate penalty for protocol violations. The reward in \Eqref{eq:calibrated-reward} modifies the original likelihood objective. Appendix~\ref{app:calibration} derives the objective and its single-sample, binary, and raw-count cases.

\Needspace{8\baselineskip}
\subsection{State-group rollout and optimization}
\label{sec:state-group}

\begin{wrapfigure}[13]{r}{0.5\textwidth}
    \vspace{-8.3pt}
    \centering
    \includegraphics[width=\linewidth]{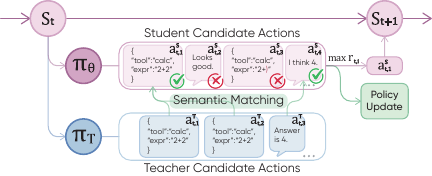}
    \caption{ActionOPD at one state. Student candidates are scored against a shared teacher multiset. One highest-reward action is committed. All candidates in retained groups contribute to training.}
    \label{fig:algorithm}
\end{wrapfigure}
\paragraph{State-group rollout.}
We use these rewards for both action selection and policy optimization (Figure~\ref{fig:algorithm}). At state $s_g$, group $g$ contains $N=4$ responses indexed by $i$, sampled as $y_{g,i}\sim\pi_{\theta_{\mathrm{old}}}(\cdot\mid s_g)$ with frozen $\theta_{\mathrm{old}}$. All candidates are scored against the same teacher multiset, yielding $r_{g,i}=r_\lambda(s_g,y_{g,i})$. We execute only the parsed action in $y_{g,i^\star}$, where $i^\star\in\argmax_i r_{g,i}$, but train on all responses in retained groups. Repeated selection produces one trajectory per task with a teacher-guided visited-state distribution, rather than the ordinary single-sample rollout distribution.

By repeatedly committing high-reward actions, state-group rollout can explore trajectories less likely under ordinary student rollouts, while introducing a degree of off-policy state visitation. Same-state candidate groups also provide local reward comparisons for policy optimization. We use $N=4$ to balance teacher-guided exploration and deviation from the student's rollout distribution.

\paragraph{Policy optimization.}
Motivated by the REINFORCE-form gradient in \Eqref{eq:action-policy-gradient}, we use group-relative advantages and a PPO-style clipped surrogate~\citep{shao2024deepseekmath,schulman2017proximal}. Groups without supervision or with identical rewards are excluded from optimization. Equal-reward groups may still advance the environment. Within each retained group, the mean reward provides a local baseline for all $N$ candidates, including invalid actions with reward $-1$. We normalize the centered rewards by a shared batch-wide standard deviation:
\begin{equation}
    \bar r_g=\frac{1}{N}\sum_{j=1}^{N}r_{g,j},
    \qquad
    \widetilde A_{g,i}=r_{g,i}-\bar r_g,
    \qquad
    A_{g,i}=\frac{\widetilde A_{g,i}}
    {\sigma+\epsilon}.
    \label{eq:group-advantage}
\end{equation}
Here $\sigma$ is the population standard deviation of centered advantages over all candidates in the retained groups, and $\epsilon>0$ stabilizes division. This shared normalization preserves relative advantage magnitudes across different state groups.

Let $\mathcal T_+$ index active generated tokens $(g,i,\ell)$ in retained groups. To compare the updated and rollout policies, define the ratio of temperature-scaled token probabilities
\begin{equation}
    \rho_{g,i,\ell}(\theta)
    =\frac{\pi_\theta(y_{g,i,\ell}\mid s_g,y_{g,i,<\ell})}
    {\pi_{\theta_{\mathrm{old}}}(y_{g,i,\ell}\mid s_g,y_{g,i,<\ell})}.
    \label{eq:token-ratio}
\end{equation}
Following the token-level expansion in Section~\ref{sec:action-likelihood}, $A_{g,i}$ weights every active token in $y_{g,i}$. PPO clipping discourages excessively large changes in token probabilities. With clipping width $\eta>0$, we maximize the token-mean surrogate
\begin{equation}
\begin{aligned}
    J_{\mathrm{clip}}(\theta)
    =\frac{1}{|\mathcal T_+|}
    \sum_{(g,i,\ell)\in\mathcal T_+}
    \min\!\bigl\{
    \rho_{g,i,\ell}(\theta)A_{g,i},\,
    \operatorname{clip}(\rho_{g,i,\ell}(\theta),1-\eta,1+\eta)A_{g,i}
    \bigr\}.
\end{aligned}
\label{eq:ppo-surrogate}
\end{equation}
With the frequency bonus, differing match counts can yield nonzero advantages even when all candidates match a teacher sample. This allows optimization to distinguish among teacher-consistent responses. These updates are not unbiased estimates of the gradient in \Eqref{eq:action-policy-gradient} after group-relative normalization, masking, and clipping. Appendix~\ref{app:optimization} details dual clipping and implementation.

\section{Experiments}
\label{sec:experiments}

We evaluate ActionOPD's cross-benchmark generalization, student--teacher scaling, and performance relative to other distillation methods. We then use ablations and case studies to examine its components and decision-making behavior, and explore Self-ActionOPD with privileged teacher context.

\subsection{Cross-benchmark transfer setup}
\label{sec:experimental-design}

\paragraph{Training data.}
We trained on Agent World Model (AWM; \citealp{wang2026awm}) and EnvScaler~\citep{song2026envscaler}. AWM provides executable tools and persistent backend state; EnvScaler adds a simulated user to the interaction loop. The curated pool comprised \textbf{5,942 AWM tasks from 794 environments} and \textbf{755 EnvScaler tasks from 49 environments}: \textbf{6,697 tasks} in total. Screening assessed environment integrity and static task feasibility, without requiring successful expert rollouts (Appendix~\ref{app:experimental-data}).

\paragraph{Models and training.}
We paired Qwen3-4B and Qwen3-8B students~\citep{yang2025qwen3} with Qwen3.7-Flash~\citep{qwen2026flash} and DeepSeek-V4-Flash~\citep{deepseek2026v4flash} action teachers in a $2\times2$ design. Each configuration used 100 training steps and 64 task starts per step (59 AWM, 5 EnvScaler), with $N=4$ student candidates, $K=3$ teacher samples per state, and frequency bonus scale $\lambda=0.5$. Task starts could revisit tasks in the curated pool. We report the final checkpoints, with Base denoting the corresponding student before ActionOPD training. Appendix~\ref{app:training-configurations} details settings and auxiliary models.

\paragraph{Evaluation.}
We evaluated transfer on \textbf{BFCL Multi-turn}~\citep{patil2025bfcl}, \textbf{ToolSandbox}'s verified-local-only subset~\citep{lu2025toolsandbox}, \textbf{AppWorld} Normal and Challenge~\citep{trivedi2024appworld}, and \textbf{$\tau^2$-bench} Airline, Retail, and Telecom~\citep{barres2025tau2}. These tasks were excluded from AWM/EnvScaler training. Students used greedy decoding without teacher-guided action selection. Appendix~\ref{app:evaluation-protocol} details coverage, trials, and interaction protocols.

\paragraph{Metrics.}
We report BFCL accuracy, ToolSandbox similarity (not binary success), AppWorld task/scenario goal completion (TGC/SGC), and Tau $\mathrm{pass}^{1}/\mathrm{pass}^{4}$: single-trial success and four-trial all-success reliability, not best-of-four success~\citep{yao2025taubench}. The \emph{average} in Table~\ref{tab:main-results} weights the four benchmarks equally, using mean Normal/Challenge TGC for AppWorld and mean domain $\mathrm{pass}^{1}$ for Tau. Appendix~\ref{app:score-definitions} gives the estimators.

\begin{table}[t]
    \centering
    \caption{Cross-benchmark transfer after 100 ActionOPD steps (\%). $+$ denotes training with the named teacher. AppWorld: task/scenario success on Normal/Challenge; scenario success requires all task variants to succeed. Tau: $\mathrm{pass}^{1}/\mathrm{pass}^{4}$. Average weights four benchmarks equally. Bold/underlined: best/second-best distinct scores per student size, including Base; ties share marks. Rankings and green average gains over Base (percentage points) use unrounded scores. Generation settings differ by role (Table~\ref{tab:model-generation}).}
    \label{tab:main-results}
    \small
    \setlength{\tabcolsep}{1.95pt}
    \renewcommand{\arraystretch}{1.3}
    \begin{tabular}{lcccccccc}
        \toprule
        \multicolumn{1}{c}{\multirow{2}{*}{\textbf{Model}}} & \multicolumn{1}{c}{\multirow{2}{*}{\textbf{BFCL}}} & \multicolumn{1}{c}{\multirow{2}{*}{\shortstack{\textbf{Tool}\\\textbf{Sandbox}}}} & \multicolumn{2}{c}{\textbf{AppWorld}} & \multicolumn{3}{c}{\textbf{$\tau^2$-Bench}} & \multirow{2}{*}{\textbf{Average}}\\
        \cmidrule(lr){4-5}\cmidrule(lr){6-8}
        & & & Normal & Challenge & Airline & Retail & Telecom & \\
        \midrule
        \rowcolor{teachergroup}
        \multicolumn{9}{l}{\textbf{Direct Teacher Evaluation}}\\
        \qwenmark Qwen3.7-Flash & 58.0 & 76.2 & 70.2\scoresep 55.4 & 36.5\scoresep 23.0 & 78.0\scoresep 54.0 & 55.9\scoresep 18.4 & 90.6\scoresep 71.9 & \avgcell{65.6}{}\\
        \deepseekmark DeepSeek-V4-Flash & 57.0 & 72.3 & 84.5\scoresep 75.0 & 74.8\scoresep 54.7 & 77.0\scoresep 52.0 & 51.8\scoresep 15.8 & 98.9\scoresep 95.6 & \avgcell{71.2}{}\\
        \midrule
        \rowcolor{studentfour!18}
        \multicolumn{9}{l}{\qwenmark\textbf{Qwen3-4B Student}}\\
        Base & 34.3 & 55.5 & 9.5\scoresep 3.6 & 2.6\scoresep \underline{0.7} & \underline{34.5}\scoresep 8.0 & 26.8\scoresep 7.9 & 20.6\scoresep 5.3 & \avgcell{30.8}{}\\
        + \qwenmark Qwen3.7-Flash & \underline{42.3} & \underline{61.8} & \textbf{14.3}\scoresep \textbf{8.9} & \underline{4.8}\scoresep \underline{0.7} & 33.5\scoresep \underline{12.0} & \underline{34.4}\scoresep \textbf{9.6} & \underline{22.4}\scoresep \underline{7.9} & \avgcell{\underline{35.9}}{5.1}\\
        + \deepseekmark DeepSeek-V4-Flash & \textbf{44.1} & \textbf{63.3} & \underline{13.1}\scoresep \underline{5.4} & \textbf{6.5}\scoresep \textbf{2.2} & \textbf{39.0}\scoresep \textbf{20.0} & \textbf{37.1}\scoresep \underline{8.8} & \textbf{24.1}\scoresep \textbf{16.7} & \avgcell{\textbf{37.7}}{6.9}\\
        \addlinespace[4pt]
        \rowcolor{studenteight!18}
        \multicolumn{9}{l}{\qwenmark\textbf{Qwen3-8B Student}}\\
        Base & 39.9 & \underline{57.7} & 15.5\scoresep 5.4 & \underline{4.6}\scoresep 0.7 & 35.0\scoresep 8.0 & \underline{38.2}\scoresep \underline{14.0} & 31.8\scoresep \textbf{14.9} & \avgcell{35.6}{}\\
        + \qwenmark Qwen3.7-Flash & \underline{45.0} & \textbf{62.5} & \underline{16.7}\scoresep \underline{8.9} & \textbf{7.7}\scoresep \underline{1.4} & \underline{36.5}\scoresep \textbf{22.0} & 37.7\scoresep 10.5 & \textbf{34.2}\scoresep \textbf{14.9} & \avgcell{\underline{39.0}}{3.3}\\
        + \deepseekmark DeepSeek-V4-Flash & \textbf{45.5} & \textbf{62.5} & \textbf{17.9}\scoresep \textbf{10.7} & \textbf{7.7}\scoresep \textbf{2.9} & \textbf{44.0}\scoresep \underline{18.0} & \textbf{40.1}\scoresep \textbf{15.8} & \underline{33.8}\scoresep \underline{12.3} & \avgcell{\textbf{40.0}}{4.4}\\
        \bottomrule
    \end{tabular}
\end{table}

\subsection{Main results}
\label{sec:main-results}

Table~\ref{tab:main-results} compares distilled students with base models and directly evaluated teachers.

\paragraph{Cross-benchmark generalization.}
After training on AWM and EnvScaler, all four configurations improved the four-benchmark average over their base models by 3.3--6.9 percentage points. Gains extended across BFCL accuracy, ToolSandbox similarity, and AppWorld task completion. These gains on benchmarks excluded from training demonstrate that action-level distillation transfers beyond the training environments, improving average performance across diverse agent tasks.

\paragraph{Scaling with student size.}
With either teacher, Qwen3-8B achieved a higher final average than Qwen3-4B under the same task-start and training-step budgets. We used the same candidate count, teacher sample count, and reward formulation for both students. This pattern supports empirical scaling across the two tested sizes, with larger students yielding stronger distilled agents.

\paragraph{Scaling with teacher capability.}
DeepSeek-V4-Flash achieved a higher direct-evaluation average than Qwen3.7-Flash and yielded higher student averages at both sizes (Figure~\ref{fig:teaser}), supporting empirical scaling with teacher capability. The DeepSeek-to-Qwen results also demonstrate effective cross-family distillation through an action-level interface that requires neither a shared model family nor tokenizer alignment. This allows teacher selection to prioritize decision quality rather than compatibility with the student's token space.

\subsection{Comparison with distillation baselines}
\label{sec:distillation-comparison}

We compared ActionOPD, offline distillation (KD), and sampled-token reverse-KL OPD using a Qwen3-4B student, a Qwen3-32B teacher, and the same 178 Tau training tasks. All methods used 24 task starts per step---eight per domain---and were evaluated at step 50. KD collected four independent teacher trajectories per task start and fine-tuned on all reasoning and action/message tokens from successful trajectories. OPD collected four student trajectories per task start and used teacher log-probabilities to supervise sampled tokens. ActionOPD used state-group rollout, sampling four candidates at each visited state while advancing one trajectory per task start. It used $K=3$ teacher samples per state and $\lambda=0.5$ (Appendix~\ref{app:distillation-baselines}).

\begin{table}[t]
    \centering
    \caption{Tau distillation comparison: $\mathrm{pass}^{1}/\mathrm{pass}^{4}$ (\%). Qwen3-4B students and a Qwen3-32B teacher. Trained students are evaluated at step 50. Four trials per task. Average equally weights domains within each split. ActionOPD uses the same run as in Table~\ref{tab:ablations}. Bold/underlined: best/second-best distinct scores (unrounded); teacher excluded.}
    \label{tab:distillation-comparison}
    \small
    \setlength{\tabcolsep}{1.8pt}
    \renewcommand{\arraystretch}{1.4}
    \begin{tabularx}{\linewidth}{l*{8}{>{\centering\arraybackslash}X}}
        \toprule
        \multicolumn{1}{c}{\multirow{2}{*}{\textbf{Model}}} & \multicolumn{2}{c}{\textbf{Airline}} & \multicolumn{2}{c}{\textbf{Retail}} & \multicolumn{2}{c}{\textbf{Telecom}} & \multicolumn{2}{c}{\textbf{Average}}\\
        \cmidrule(lr){2-3}\cmidrule(lr){4-5}\cmidrule(lr){6-7}\cmidrule(lr){8-9}
        & Train (30) & Test (20) & Train (74) & Test (40) & Train (74) & Test (40) & Train & Test\\
        \midrule
        Qwen3-32B & 49.2\scoresep 16.7 & 43.8\scoresep 15.0 & 46.0\scoresep 16.2 & 46.9\scoresep 15.0 & 25.7\scoresep 13.5 & 41.9\scoresep 30.0 & 40.3\scoresep 15.5 & 44.2\scoresep 20.0\\
        \midrule
        Base & 35.0\scoresep 10.0 & 33.8\scoresep 5.0 & 26.7\scoresep 6.8 & 26.9\scoresep \underline{10.0} & 17.2\scoresep 4.1 & 26.9\scoresep 7.5 & 26.3\scoresep 6.9 & 29.2\scoresep 7.5\\
        KD & 38.3\scoresep \textbf{20.0} & 37.5\scoresep \textbf{20.0} & 37.5\scoresep \textbf{13.5} & 33.8\scoresep \textbf{12.5} & \underline{25.3}\scoresep \textbf{13.5} & \textbf{41.3}\scoresep \textbf{25.0} & 33.7\scoresep \textbf{15.7} & 37.5\scoresep \textbf{19.2}\\
        OPD & \underline{40.0}\scoresep 13.3 & \textbf{41.3}\scoresep \textbf{20.0} & \underline{40.2}\scoresep \underline{10.8} & \underline{37.5}\scoresep \underline{10.0} & 22.6\scoresep \underline{10.8} & 40.0\scoresep \underline{22.5} & \underline{34.3}\scoresep 11.7 & \underline{39.6}\scoresep \underline{17.5}\\
        \rowcolor{studentfour!18}
        \textbf{ActionOPD} & \textbf{42.5}\scoresep \underline{16.7} & \underline{40.0}\scoresep \underline{15.0} & \textbf{42.9}\scoresep \underline{10.8} & \textbf{39.4}\scoresep \textbf{12.5} & \textbf{28.4}\scoresep \textbf{13.5} & \underline{40.6}\scoresep 15.0 & \textbf{37.9}\scoresep \underline{13.7} & \textbf{40.0}\scoresep 14.2\\
        \bottomrule
    \end{tabularx}
\end{table}

\begin{table}[t]
    \centering
    \caption{Tau ablations: $\mathrm{pass}^{1}/\mathrm{pass}^{4}$ (\%). Qwen3-4B student, Qwen3-32B teacher; trained variants at step 50. Four trials per task. Parentheses: task counts. Average weights domains equally per split. Bold/underlined: best/second-best distinct unrounded scores, including Base; ties share marks.}
    \label{tab:ablations}
    \small
    \setlength{\tabcolsep}{2.4pt}
    \renewcommand{\arraystretch}{1.4}
    \begin{tabularx}{\linewidth}{l*{8}{>{\centering\arraybackslash}X}}
        \toprule
        \multicolumn{1}{c}{\multirow{2}{*}{\textbf{Model}}} & \multicolumn{2}{c}{\textbf{Airline}} & \multicolumn{2}{c}{\textbf{Retail}} & \multicolumn{2}{c}{\textbf{Telecom}} & \multicolumn{2}{c}{\textbf{Average}}\\
        \cmidrule(lr){2-3}\cmidrule(lr){4-5}\cmidrule(lr){6-7}\cmidrule(lr){8-9}
        & Train (30) & Test (20) & Train (74) & Test (40) & Train (74) & Test (40) & Train & Test\\
        \midrule
        Base & 35.0\scoresep \underline{10.0} & \underline{33.8}\scoresep 5.0 & 26.7\scoresep 6.8 & 26.9\scoresep 10.0 & 17.2\scoresep 4.1 & 26.9\scoresep 7.5 & 26.3\scoresep 6.9 & 29.2\scoresep 7.5\\
        Validity Reward & \textbf{42.5}\scoresep \textbf{16.7} & 30.0\scoresep \underline{10.0} & 23.7\scoresep 8.1 & 26.3\scoresep 10.0 & \underline{25.3}\scoresep \textbf{14.9} & 36.3\scoresep 10.0 & 30.5\scoresep \underline{13.2} & 30.8\scoresep 10.0\\
        Binary Reward & \underline{35.8}\scoresep \underline{10.0} & \underline{33.8}\scoresep \underline{10.0} & 35.8\scoresep 8.1 & 36.3\scoresep \textbf{17.5} & \textbf{28.4}\scoresep \underline{13.5} & \underline{40.6}\scoresep \textbf{15.0} & \underline{33.3}\scoresep 10.5 & 36.9\scoresep \textbf{14.2}\\
        Random Commit & 34.2\scoresep 3.3 & 27.5\scoresep \underline{10.0} & \underline{41.6}\scoresep \textbf{16.2} & \textbf{40.6}\scoresep \underline{15.0} & 24.0\scoresep 8.1 & \textbf{44.4}\scoresep \underline{12.5} & 33.2\scoresep 9.2 & \underline{37.5}\scoresep \underline{12.5}\\
        \rowcolor{studentfour!18}
        \textbf{ActionOPD} & \textbf{42.5}\scoresep \textbf{16.7} & \textbf{40.0}\scoresep \textbf{15.0} & \textbf{42.9}\scoresep \underline{10.8} & \underline{39.4}\scoresep 12.5 & \textbf{28.4}\scoresep \underline{13.5} & \underline{40.6}\scoresep \textbf{15.0} & \textbf{37.9}\scoresep \textbf{13.7} & \textbf{40.0}\scoresep \textbf{14.2}\\
        \bottomrule
    \end{tabularx}
\end{table}

KD achieved the highest test average $\mathrm{pass}^{4}$ (19.2), indicating greater repeated-success reliability on limited tasks and states. Conversely, OPD and ActionOPD achieved higher test average $\mathrm{pass}^{1}$ than KD. This pattern demonstrates the advantages of on-policy supervision: reducing mismatch between teacher demonstrations and student-encountered states, and improving performance across a broader range of student tasks. ActionOPD achieved higher training-set averages than OPD on both metrics and exceeded OPD's test average $\mathrm{pass}^{1}$ by 0.4 percentage points, although its test average $\mathrm{pass}^{4}$ was lower. While these results do not establish uniform superiority, ActionOPD offers a competitive alternative with a more flexible teacher interface. Its generation-only interface requires neither teacher token log-probabilities nor tokenizer alignment, and does not require a shared model family. The cross-family transfer demonstrated in Section~\ref{sec:main-results} complements this same-family comparison.

\subsection{Ablation studies}
\label{sec:ablations}

Ablations shared the settings in Section~\ref{sec:distillation-comparison}, with $N=4$ and $K=3$. Validity Reward replaced the semantic training reward with action validity, while retaining semantic scores for action selection. Binary Reward removed the frequency bonus from \Eqref{eq:calibrated-reward}, using binary agreement for both training and action selection. Random Commit retained ActionOPD's reward but selected uniformly from all candidates, including invalid or zero-reward actions. Appendix~\ref{app:ablation-definitions} details the variants and their effects on trainable state groups.

Table~\ref{tab:ablations} reports results on the training set and 100 held-out tasks. ActionOPD outperforms all three variants in average single-trial success on both splits, supporting the combined design. The largest drop occurs with Validity Reward, which assigns identical rewards to all valid candidates and thus provides no learning signal in all-valid groups. Semantic rewards can instead distinguish these candidates by their agreement with teacher actions, highlighting the value of supervising action meaning beyond protocol compliance.

\begin{table}[t]
    \centering
    \caption{Selected positive cases: Qwen3-4B before/after 100 ActionOPD steps with DeepSeek-V4-Flash. Fractions: successes in the first four trials (Base $\rightarrow$ ActionOPD). Each pair contrasts Base failure with ActionOPD success on the same task and trial. Outcomes denote evaluation results, not training rewards.}
    \label{tab:decision-patterns}
    \small
    \setlength{\fboxsep}{1pt}
    \setlength{\tabcolsep}{4pt}
    \renewcommand{\arraystretch}{1.1}
    \begin{tabularx}{\linewidth}{>{\hsize=0.90\hsize\linewidth=\hsize\raggedright\arraybackslash}X>{\hsize=0.96\hsize\linewidth=\hsize\raggedright\arraybackslash}X>{\hsize=1.14\hsize\linewidth=\hsize\raggedright\arraybackslash}X}
        \toprule
        \rowcolor{studentfour!18}
        \textbf{Pattern / Task Context} & \textbf{Qwen3-4B Base} & \textbf{Qwen3-4B + ActionOPD}\\
        \midrule
        \textbf{Following Policy Constraints}
        \par\vspace{2pt} Airline ($2/4\rightarrow4/4$). The user requests a flight change, but basic economy bookings cannot be modified.
        & Reads that the booking is basic economy, yet modifies the flight through the API, violating the policy.
        \par\vspace{2pt} \textbf{Outcome: \colorbox{red!10}{\textcolor{red!55!black}{Failure}}.}
        & Explains the restriction and calls the transfer tool when the user requests escalation. Success means keeping the booking unchanged, not rebooking.
        \par\vspace{2pt} Prohibited modifications: $2/4\rightarrow0/4$.
        \par\vspace{2pt} \textbf{Outcome: \colorbox{aopdgain!15}{\textcolor{aopdgain}{Success}}.}\\
        \midrule
        \textbf{Tracking the Correct Target}
        \par\vspace{2pt} Retail ($0/4\rightarrow3/4$). Change the target order's address and speaker. Retrieve the new address in New York from another order.
        & Treats the source order's New York address as if it already belonged to the target. Changes only the speaker. The target's address remains in Los Angeles.
        \par\vspace{2pt} \textbf{Outcome: \colorbox{red!10}{\textcolor{red!55!black}{Failure}}.}
        & Retrieves the source address, then updates the target's address followed by its speaker. The returned order contains both the New York address and the requested green speaker.
        \par\vspace{2pt} \textbf{Outcome: \colorbox{aopdgain!15}{\textcolor{aopdgain}{Success}}.}\\
        \midrule
        \textbf{Completing and Verifying Actions}
        \par\vspace{2pt} Telecom ($0/4\rightarrow2/4$). Restore service after an overdue-bill suspension and an unseated SIM. Payment alone does not reactivate the line.
        & Promises restoration after payment but never calls \texttt{resume\_line}. The user pays and reboots, but service remains disconnected and the agent transfers the case.
        \par\vspace{2pt} \textbf{Outcome: \colorbox{red!10}{\textcolor{red!55!black}{Failure}}.}
        & Recovers from an invalid tool call, resumes the line with \texttt{resume\_line}, and guides the \emph{user} to reseat the SIM. User-side tool output confirms connection, and the evaluator verifies that the debt is cleared.
        \par\vspace{2pt} \textbf{Outcome: \colorbox{aopdgain!15}{\textcolor{aopdgain}{Success}}.}\\
        \bottomrule
    \end{tabularx}
\end{table}

\begin{table}[t]
    \centering
    \caption{Self-ActionOPD on Tau: $\mathrm{pass}^{1}/\mathrm{pass}^{4}$ (\%), using Qwen3-4B. Privileged Teacher is the initial Base model given privileged context. Self-ActionOPD is evaluated at step 50 using public context only. Four trials per task. Average equally weights domains within each split. Bold/underlined: best/second-best of Base and Self-ActionOPD; ties share marks; privileged reference excluded.}
    \label{tab:self-aopd}
    \small
    \setlength{\tabcolsep}{1.8pt}
    \renewcommand{\arraystretch}{1.4}
    \begin{tabularx}{\linewidth}{l*{8}{>{\centering\arraybackslash}X}}
        \toprule
        \multicolumn{1}{c}{\multirow{2}{*}{\textbf{Model}}} & \multicolumn{2}{c}{\textbf{Airline}} & \multicolumn{2}{c}{\textbf{Retail}} & \multicolumn{2}{c}{\textbf{Telecom}} & \multicolumn{2}{c}{\textbf{Average}}\\
        \cmidrule(lr){2-3}\cmidrule(lr){4-5}\cmidrule(lr){6-7}\cmidrule(lr){8-9}
        & Train (30) & Test (20) & Train (74) & Test (40) & Train (74) & Test (40) & Train & Test\\
        \midrule
        Privileged Teacher & 51.7\scoresep 33.3 & 35.0\scoresep 15.0 & 38.2\scoresep 6.8 & 38.1\scoresep 17.5 & 20.6\scoresep 8.1 & 37.5\scoresep 20.0 & 36.8\scoresep 16.1 & 36.9\scoresep 17.5\\
        \midrule
        Base & \underline{35.0}\scoresep \underline{10.0} & \underline{33.8}\scoresep \underline{5.0} & \underline{26.7}\scoresep \underline{6.8} & \underline{26.9}\scoresep \textbf{10.0} & \underline{17.2}\scoresep \underline{4.1} & \underline{26.9}\scoresep \underline{7.5} & \underline{26.3}\scoresep \underline{6.9} & \underline{29.2}\scoresep \underline{7.5}\\
        \rowcolor{studentfour!18}
        \textbf{Self-ActionOPD} & \textbf{44.2}\scoresep \textbf{26.7} & \textbf{35.0}\scoresep \textbf{10.0} & \textbf{35.1}\scoresep \textbf{8.1} & \textbf{35.6}\scoresep \textbf{10.0} & \textbf{19.3}\scoresep \textbf{5.4} & \textbf{31.3}\scoresep \textbf{10.0} & \textbf{32.9}\scoresep \textbf{13.4} & \textbf{34.0}\scoresep \textbf{10.0}\\
        \bottomrule
    \end{tabularx}
\end{table}

\subsection{Pattern analysis}
\label{sec:pattern-analysis}

We compared Qwen3-4B Base with the same student after 100 ActionOPD steps using DeepSeek-V4-Flash as teacher. Each selected case has at least two additional successes after training over the first four trials. Table~\ref{tab:decision-patterns} summarizes one Base-failure/ActionOPD-success pair matched by task and trial index in each Tau domain, using replies, tool calls, and environment results.

In these cases, the student respected policy constraints, distinguished information sources from operation targets, and followed through on promised actions. These changes highlight the value of ActionOPD's semantic action supervision for task-appropriate behavior.

\subsection{Self-ActionOPD}
\label{sec:self-aopd}

We further examine whether a model can provide useful action-level supervision for itself when given privileged task information. Self-ActionOPD shared current Qwen3-4B weights between teacher and student, as in OPSD~\citep{zhao2026selfdistilled}, but independently sampled their responses. Teacher responses were refreshed from the updated shared model at each collection step and held fixed during subsequent policy optimization. The student received only public interaction context, whereas the teacher additionally received action guidance, summaries of private user goals and constraints, and reference actions and target outcomes. Training used action-agreement rewards on student-generated candidates only. Appendix~\ref{app:self-aopd} provides implementation details.

After 50 steps, Self-ActionOPD improved test average $\mathrm{pass}^{1}/\mathrm{pass}^{4}$ over Base by 4.8/2.5 percentage points (Table~\ref{tab:self-aopd}). Single-trial success also improved across all three domains. Unlike the privileged-teacher reference, the trained student was evaluated using only public context, showing that the student retained benefits from the privileged guidance after training. Together, these results support the feasibility of using a same-model teacher given privileged context for action-level distillation.

\section{Related work}
\label{sec:related-work}

\paragraph{Distillation and self-distillation.}
Classical distillation transfers predictive distributions~\citep{hinton2015distilling}, while sequence distillation imitates generated outputs~\citep{kim2016sequence}. Offline imitation uses token-level losses on fixed demonstrations that may not cover student-induced prefixes. GKD and MiniLLM instead supervise student-generated sequences through token-level divergences~\citep{agarwal2024onpolicy,gu2024minillm}. These objectives require teacher distributions or sampled-token log-probabilities. Guidance can become unreliable under prefix drift or special-token mismatch~\citep{fu2026revisiting}. Cross-tokenizer OPD transfers probabilistic supervision through explicit alignment~\citep{niu2026breaking,wang2026byteprefix}. Privileged self-distillation augments teacher context: On-Policy Self-Distillation (OPSD) minimizes token-level divergence on student rollouts using a shared-weight teacher conditioned on reference solutions~\citep{zhao2026selfdistilled}. SDPO distills feedback-conditioned predictions~\citep{hubotter2026reinforcement}, while privileged-information distillation studies shared-parameter training, including on-policy reverse KL~\citep{penaloza2026privileged}. Self-ActionOPD instead uses sampled-action agreement.

\paragraph{Black-box and semantic distillation.}
GAD jointly trains a discriminator to reward teacher-like responses~\citep{ye2025blackbox}, whereas ROPD evaluates responses against rubrics derived from teacher--student contrasts~\citep{fang2026rubric}. On-policy Verbal Distillation (OVD) uses verbal teacher scores to evaluate reasoning quality at step and trajectory levels~\citep{xiong2026ovd}. OmniOPD compares selected student reasoning chunks with teacher continuations sampled at student prefixes~\citep{zhou2026omniopd}. MENTOR rewards trajectory-wide tool-set agreement, final correctness, and execution validity~\citep{choi2025mentor}. Action Semantics Learning (ASL) derives a semantic-agreement policy gradient and uses a world model to compare UI transitions induced by generated and reference actions. Its RL variant combines teacher-action agreement with environment rewards~\citep{tang2026actionsemantics}. ActionOPD instead matches immediate tool calls or communicative responses against teacher samples at the same state. Its matcher is instructed to assess agreement, not correctness. ActionOPD requires no reasoning-text matching, complete reference trajectories, or terminal rewards.

\paragraph{Interactive agent learning and process supervision.}
DAgger queries experts at learner-visited states~\citep{ross2011reduction}. SCoRe provides local teacher corrections before supervised and short-horizon RL refinement~\citep{lyu2026studentcentered}. Guided-OPD gradually withdraws teacher turns through a rollout curriculum; ATOD combines OPD and RL with annealed coefficients and turn-level weighting~\citep{li2026guided,tan2026atod}. GiGPO retrospectively groups repeated states across trajectories and derives relative advantages from returns~\citep{feng2025gigpo}. ActionOPD scores directly sampled same-state candidates against a shared teacher multiset, without estimating returns.

\section{Conclusion and limitations}
\label{sec:conclusion}

ActionOPD decouples teacher decision supervision from student token realization through semantic agreement at student-encountered states, requiring only generation access to the teacher. Across tested configurations, it improves average performance on unseen benchmarks, with higher final averages from larger students and stronger teachers. ActionOPD also achieves competitive performance against offline distillation and token-level OPD. Ablations support semantic rewards over validity rewards. Self-ActionOPD improves held-out performance using a same-model teacher with privileged context, highlighting a potential path toward agent self-improvement. Together, these findings support semantic actions as a unit of supervision for agent distillation.

Three limitations warrant attention. First, agreement does not guarantee correctness: teacher errors or suboptimal decisions can become learning targets. Second, LLM matching of communicative responses and tool-argument differences can produce false positives or negatives, misassigning rewards. Third, fixed-size teacher and student sample sets may contain few semantically matching actions when plausible decisions are dispersed, as may occur in long-horizon interactions. Increasing $K$ and $N$ could improve coverage at greater generation and matching cost, but the benefit in this regime remains untested.

\section*{AI use statement}
The authors developed the method and designed all experiments. Generative AI tools assisted with manuscript preparation, literature review, implementation, debugging, derivation checks, and result analysis, and provided feedback on methodological details. The use of language models for task screening, teacher supervision, semantic matching, and user simulation is documented in the experimental protocol. All final research decisions were made by the authors, who take responsibility for the manuscript and its findings.

\section*{Reproducibility statement}
Appendices~\ref{app:method} and~\ref{app:experimental-details} document the derivations, algorithm, data preparation, and experimental protocols. Code, data, and model checkpoints for the core results will be made available at \url{https://github.com/thu-nics/ActionOPD}.

\bibliography{iclr2027_conference}
\bibliographystyle{iclr2027_conference}

\clearpage
\appendix
\section{Derivations and implementation details}
\label{app:method}

\subsection{From teacher-action likelihood to REINFORCE}
\label{app:derivation}

\paragraph{Assumptions and induced actions.}
Let $\mathcal Y(s)$ contain student response sequences, including termination tokens. At each state, semantic action classes partition these sequences: $\phi_s(y)$ maps equivalent realizations to the same action, and protocol-invalid outputs to an invalid outcome $\bot$. The induced $\pi_\theta(a\mid s)$ is normalized over the complete action space, including $\bot$. Teacher targets from $\pi_{\mathrm T}(a\mid s)$ are valid actions. Summation over actions counts each semantic class once, not each expression of that class.

Assume a normalized, differentiable response policy on a fixed support, with a finite vocabulary and bounded response length. These conditions permit exchanging differentiation and summation. The semantic mapping $\phi_s$, teacher distribution $\pi_{\mathrm T}$, and collected-state distribution $d_{\mathrm S}$ remain fixed during each update. Teacher action samples suffice; teacher token probabilities are not needed.

\paragraph{Expanding the likelihood.}
For a teacher semantic action, all its student realizations contribute to its likelihood,
\begin{equation}
    \pi_\theta(a^{\mathrm T}\mid s)
    =\sum_{y\in\mathcal Y(s)}\pi_\theta(y\mid s)
      \mathbf 1\{\phi_s(y)=a^{\mathrm T}\}.
\end{equation}
Consequently,
\begin{equation}
\begin{aligned}
    J_{\mathrm{ActionOPD}}(\theta)
    &=\mathbb E_{s\sim d_{\mathrm S}}
      \sum_{a^{\mathrm T}}\pi_{\mathrm T}(a^{\mathrm T}\mid s)
      \sum_y\pi_\theta(y\mid s)\mathbf 1\{\phi_s(y)=a^{\mathrm T}\}\\
    &=\mathbb E_{s\sim d_{\mathrm S}}
      \sum_y\pi_\theta(y\mid s)r_{\mathrm T}(s,\phi_s(y)).
\end{aligned}
\end{equation}
Differentiating the student factor gives
\begin{equation}
\begin{aligned}
    \nabla_\theta J_{\mathrm{ActionOPD}}(\theta)
    &=\mathbb E_s\sum_y r_{\mathrm T}(s,\phi_s(y))\nabla_\theta\pi_\theta(y\mid s)\\
    &=\mathbb E_{s,y}\!\left[r_{\mathrm T}(s,\phi_s(y))
      \nabla_\theta\log\pi_\theta(y\mid s)\right]\\
    &=\mathbb E_{s,a^{\mathrm T},y}\!\left[
      \mathbf 1\{\phi_s(y)=a^{\mathrm T}\}
      \sum_{\ell=1}^{|y|}\nabla_\theta\log\pi_\theta(y_\ell\mid s,y_{<\ell})
      \right].
\end{aligned}
\label{eq:app-gradient}
\end{equation}
This score-function estimator~\citep{williams1992simple} uses semantic action agreement without evaluating the marginal $\pi_\theta(a^{\mathrm T}\mid s)$ explicitly. The likelihood and agreement objectives are identical under this semantic action definition.

An action-independent baseline $b(s)$ leaves this gradient unchanged because $\mathbb E_y[\nabla_\theta\log\pi_\theta(y\mid s)]=0$. A sample group mean instead includes each candidate's own reward. For independent candidates and an otherwise unmodified estimator, subtracting that mean scales the expected gradient by $(N-1)/N$. Whitening, masking, token normalization, and clipping further distinguish the practical surrogate from the ideal gradient.

\paragraph{Expected likelihood versus expected log-likelihood.}
Where $\pi_\theta(a^{\mathrm T}\mid s)>0$, replacing the expected likelihood in \Eqref{eq:distillation-objective} with $\mathbb E_{s,a^{\mathrm T}}[\log\pi_\theta(a^{\mathrm T}\mid s)]$ yields the following gradient:
\begin{equation}
    \mathbb E_{s,a^{\mathrm T},y}
    \left[
    \frac{\mathbf 1\{\phi_s(y)=a^{\mathrm T}\}}{\pi_\theta(a^{\mathrm T}\mid s)}
    \nabla_\theta\log\pi_\theta(y\mid s)
    \right].
\end{equation}
The inverse-likelihood factor is absent from ActionOPD. Expected agreement also need not preserve teacher modes: for disjoint actions $A,B$ with teacher probabilities $0.7,0.3$, a student placing all mass on $A$ maximizes the unregularized likelihood objective.

\Needspace{8\baselineskip}
\subsection{Objective induced by the frequency bonus}
\label{app:calibration}

Fix a state $s$, a nominal teacher sample count $K>1$, and the retained teacher-response multiset $\mathcal M=\mathcal M_{\mathrm T}(s)$ with $1\le|\mathcal M|\le K$. Write $c(y)=c(s,y)=\sum_{k=1}^{|\mathcal M|}m_s(y,y_k^{\mathrm T})$, suppressing the fixed state and multiset. Let
\begin{equation}
\begin{aligned}
    C_\theta(s;\mathcal M)
      &=\mathbb E_{y\sim\pi_\theta}[c(y)],\\
    U_\theta(s;\mathcal M)
      &=\mathbb P_{y\sim\pi_\theta}[c(y)>0],\\
    V_\theta(s)
      &=\mathbb P_{y\sim\pi_\theta}[y\text{ contains an invalid action}].
\end{aligned}
\end{equation}
Here $C_\theta$ is the expected match count, $U_\theta$ the any-match probability, and $V_\theta$ the invalid-action probability. With accurate matching, $C_\theta=\sum_k\pi_\theta(a_k^{\mathrm T}\mid s)$, but the decomposition below does not require this equality. Invalid actions have $c(y)=0$. Any-match probability does not measure coverage of distinct teacher actions. Writing $\alpha=\lambda/(K-1)$, \Eqref{eq:calibrated-reward} is exactly
\begin{equation}
    r_\lambda(s,y)
    =\alpha c(y)+(1-\alpha)\mathbf 1\{c(y)>0\}
      -\mathbf 1\{y\text{ contains an invalid action}\}.
\end{equation}
Its expected-reward objective is therefore
\begin{equation}
    J_\lambda(\theta\mid s,\mathcal M)
    =\alpha C_\theta(s;\mathcal M)
      +(1-\alpha)U_\theta(s;\mathcal M)-V_\theta(s).
    \label{eq:app-calibrated-objective}
\end{equation}
The same score-function derivation applies to $r_\lambda$. For $0\le\lambda\le K-1$, the positive terms combine expected match count and any-match probability with nonnegative weights. Beyond this range, the identity still holds but the any-match coefficient is negative. For $0<\lambda<K-1$, each additional match contributes $\alpha<1$, whereas the first match earns one. This downweights additional matches relative to the first match, but does not imply reduced variance of the normalized policy update. This objective is conditional on the retained samples and does not calibrate teacher probabilities or confidence.

At $\lambda=0$, valid actions receive binary agreement rewards. At $\lambda=K-1$, their reward is the raw match count $c$, while the invalid-action penalty remains $-1$. With $K=3,\lambda=0.5$, $\alpha=0.25$, and $[A,A,B]$, rewards are $1.25,1,0,-1$ for an $A$ match, a $B$ match, a valid non-match, and an invalid action. One retained teacher action gives unit reward for a match and objective $U_\theta-V_\theta$, as does an explicit $K=1$ configuration. An empty multiset supplies no supervision, not negative labels.

\begin{algorithm}[t]
\caption{ActionOPD}
\label{alg:aopd}
\small
\begin{algorithmic}[1]
\Require Student $\pi_\theta$, teacher $\pi_{\mathrm T}$, matcher $m_s$, tasks, $K=3$, $N=4$, $\lambda=0.5$
\For{each training iteration}
    \State Freeze $\theta_{\mathrm{old}}\gets\theta$; initialize state-group buffer $\mathcal B\gets\emptyset$
    \For{each sampled task}
        \State Reset the environment and obtain public state $s$
        \While{task is active and interaction budget remains}
            \State Set group index $g\gets|\mathcal B|+1$ and state $s_g\gets s$
            \State Sample teacher responses $\mathcal M_{\mathrm T}(s_g)=[y^{\mathrm T}_1,\ldots,y^{\mathrm T}_K]$ using $\pi_{\mathrm T}$
            \State Sample $y_{g,i}\sim\pi_{\theta_{\mathrm{old}}}(\cdot\mid s_g)$ and retain parsed actions, for $i=1,\ldots,N$
            \State Compute $c(s_g,y_{g,i})$ and $r_{g,i}=r_\lambda(s_g,y_{g,i})$ (\Eqref{eq:teacher-count}; \Eqref{eq:calibrated-reward})
            \State Add the complete state group $(s_g,\{y_{g,i},r_{g,i}\}_{i=1}^{N})$ to $\mathcal B$
            \State Select $i^\star\in\argmax_i r_{g,i}$; execute the parsed action in $y_{g,i^\star}$ and update $s$
        \EndWhile
    \EndFor
    \State Exclude groups without supervision or with equal rewards
    \State Center all $N$ rewards per retained group and whiten across the batch (\Eqref{eq:group-advantage})
    \State Broadcast advantages to generated tokens and update $\theta$ with the clipped objective (\Eqref{eq:ppo-surrogate})
\EndFor
\end{algorithmic}
\end{algorithm}

\subsection{Teacher supervision and semantic matching}
\label{app:protocol}

\paragraph{Visible information and native tools.}
In standard ActionOPD, teacher and student receive the same retained public context and tool schemas through their native chat templates. Each teacher sample contains one immediate action. Hidden simulator state, verifier code, task solutions, and private user-simulator instructions are excluded. Self-ActionOPD adds the inputs in Appendix~\ref{app:self-aopd}. Tools are exposed directly, without \texttt{list\_tools}/\texttt{call\_tool} wrappers. Multiple student tool calls violate the single-action protocol; only the teacher's first call is retained when its output contains several.

\paragraph{Teacher sampling and validation.}
The teacher multiset is fixed before student generation. Each of three requested samples permits two retries for parser/schema-invalid output; valid samples are retained without regeneration. This filters protocol validity, not task success. A nonempty partial multiset can supervise a group, retaining nominal $K=3$ in the bonus denominator. With two retained copies of $A$, a matching student response receives $1.25$ at $\lambda=0.5$. With no valid samples, collection ends without a student group or policy-failure label. Retries and partial retention differ from independent sampling in Section~\ref{sec:semantic-supervision}.

\paragraph{Caching.}
Teacher-cache keys include task/state context, tools, teacher/provider, prompt and protocol versions, decoding settings, and history-dependent state. Retained samples preserve multiplicity. Reuse does not provide fresh independent draws. Matcher-cache keys include the action pair, evidence, model, and prompt.

\paragraph{Context management.}
Native chat templates retain fixed task/system content and complete tool schemas, dropping the oldest removable exchanges until the prompt fits its token budget. The standard teacher uses the same retained public context; Self-ActionOPD has an additional teacher-only prompt allowance (Appendix~\ref{app:self-aopd}). If protected content cannot fit, the rollout terminates without truncating tool definitions or fabricating an action. Prompt construction leaves the environment's full interaction history intact.

\paragraph{Semantic matching.}
\label{app:matching}

The implemented $m_s(y,y^{\mathrm T})$ compares retained tool-call fields or message text, not reasoning. It approximates $\mathbf 1\{\phi_s(y)=\phi_s(y^{\mathrm T})\}$. Approximation errors can produce inconsistent judgments for semantically equivalent responses. Thus counts and rewards depend on sampled responses, not only their semantic classes. For fixed $y$ and a complete multiset of $K$ independent teacher draws, $c/K$ is unbiased for $\mathbb E_{y^{\mathrm T}}[m_s(y,y^{\mathrm T})]$. Accurate matching recovers $r_{\mathrm T}(s,\phi_s(y))$. With any fixed matcher, the score-function identity still applies to its expected reward, but errors can make that objective differ from $J_{\mathrm{ActionOPD}}$.

The matcher has fixed parameters and prompts. It is instructed to assess agreement in the immediate operation or communicative act, including facts, identifiers, numbers, negation, commitments, and exact-copy requirements. Shared usefulness or outcomes do not establish equivalence. Ambiguous judgments are negative; API or response-format failures mean unavailable supervision, not negative matches.

Tool calls must name the same tool. Canonical checks ignore dictionary-key order and normalize integral floating-point values, but distinguish booleans from numbers. Canonicalization generally preserves distinctions between missing and null values, as well as list order and multiplicity; tool-source evidence can justify specific defaults or normalization. Remaining differences are judged using the tool schema, both argument objects, differing paths, public context, and tool-implementation evidence. Every teacher sample contributes a separate binary match, even when identical pairs reuse cached judgments. No transitive closure is applied.

Tool-implementation evidence is available to the \emph{matcher}, not the teacher policy. It supports argument interpretation, not inspection of hidden database contents, task answers, or action quality. Two calls do not match merely because both fail or produce no change. Missing required source evidence makes supervision unavailable. Message matching receives both messages, public context, and tools, but no hidden solution. Normalized exact matches bypass the LLM. Cross-kind pairs do not match, but tool, message, and invalid candidates remain in the same group.

\Needspace{8\baselineskip}
\subsection{Optimization and execution details}
\label{app:optimization}

\paragraph{Active groups and tokens.}
Training retains complete groups with available supervision and a reward range greater than $10^{-8}$. All $N$ responses in each retained group contribute to reward centering, including invalid actions with reward $-1$. Centered rewards are divided by their population standard deviation across all retained groups, with $\epsilon=10^{-8}$. No additional mean subtraction is needed. A positive-reward action can have negative advantage, and an all-matching group remains informative if match counts differ. An empty active batch skips optimization.

Advantages apply to all active student-generated tokens, including reasoning, but not to prompts, tool observations, or user-simulator outputs. The actor uses active-token mean aggregation. Padding rows serve computational alignment and contribute neither statistics nor gradients. Compact execution skips masked training rows while retaining rollout records and metrics, without introducing within-group candidate selection. Unselected responses in retained groups still contribute to training.

\paragraph{Clipped optimization.}
Let $h(\rho,A)=\min\{\rho A,\operatorname{clip}(\rho,1-\eta,1+\eta)A\}$. The implementation additionally uses dual clipping~\citep{ye2020mastering},
\begin{equation}
    h_{\mathrm{dual}}(\rho,A)=
    \begin{cases}
        \max\{h(\rho,A),c_{\mathrm{dual}}A\}, & A<0,\\
        h(\rho,A), & A\ge0,
    \end{cases}
    \qquad \eta=0.2,\quad c_{\mathrm{dual}}=3.
\end{equation}
The actor maximizes the mean of $h_{\mathrm{dual}}$ over active tokens. Teacher and matcher outputs, old-policy scores, and advantages remain fixed during the update. No reference-policy KL penalty or entropy bonus is added.

\paragraph{Sampling and asynchronous collection.}
Table~\ref{tab:model-generation} specifies student and teacher generation settings. In \Eqref{eq:token-ratio}, old and updated $\pi$ scores use the full softmax at the rollout temperature, without top-$p$/top-$k$ truncation. Their ratio does not correct for truncation during candidate sampling.

Workers overlap teacher sampling and tool/API waiting with generation at other ready states. Within each group, teacher samples are fixed before student generation. Collection uses one behavior-policy snapshot, followed by optimization. Algorithm~\ref{alg:aopd} specifies these dependencies without requiring sequential workers.

\section{Experimental details}
\label{app:experimental-details}

\subsection{Training data and task selection}
\label{app:experimental-data}

Table~\ref{tab:data-funnel} records the transfer-training pools. AWM context selection retained tasks whose initial native prompt, including tool schemas, contained at most 16,000 tokens under the Qwen3-4B tokenizer. This selection cutoff is distinct from the training context limit. A deterministic audit checked scenario, database, and verifier integrity. Static feasibility screening then assessed each task using its tools, implementation, initial data, and verification evidence. It tested whether the environment supported the requested goal and success conditions, not whether a particular policy could succeed. EnvScaler used the same feasibility criterion after deterministic checks. Judge confidence was diagnostic, not an admission threshold.

Static feasibility screening for both data sources used DeepSeek-V4-Flash. Manifests link retained tasks to source data and screening records. Admission did not require expert rollout success. Static screening does not rule out runtime defects.

Each 100-step transfer run contains 5,900 AWM and 500 EnvScaler task starts, potentially revisiting tasks from the 6,697-task pool. Tau comparisons, ablations, and Self-ActionOPD use the official 178-task training and 100-task test splits.

\begin{table}[H]
    \centering
    \caption{Task-selection funnel for the pools used in this study. Task counts denote unique tasks remaining after each stage, not rollout attempts. The final row reports environment coverage. Context selection is an AWM-specific stage. The dash denotes that no separate corresponding stage was used for EnvScaler.}
    \label{tab:data-funnel}
    \small
    \setlength{\tabcolsep}{4pt}
    \renewcommand{\arraystretch}{1.15}
    \begin{tabular*}{\linewidth}{@{}l@{\extracolsep{\fill}}rr@{}}
        \toprule
        \textbf{Stage} & \textbf{AWM} & \textbf{EnvScaler}\\
        \midrule
        Source tasks & 10,000 & 2,550\\
        Context-eligible & 9,380 & ---\\
        Deterministic integrity pass & 7,562 & 2,495\\
        Static feasibility: healthy & 5,942 & 755\\
        Environments represented in the final pool & 794 & 49\\
        \bottomrule
    \end{tabular*}
\end{table}

\subsection{Training configurations and auxiliary models}
\label{app:training-configurations}

Tables~\ref{tab:training-summary} and~\ref{tab:model-generation} summarize training and generation settings. Teacher/matcher inputs follow Appendix~\ref{app:protocol}. User simulators generate interaction turns, not distillation targets. Our implementation extends \texttt{verl-agent}~\citep{feng2025gigpo}, built on veRL/HybridFlow~\citep{sheng2025hybridflow}.

\begin{table}[t]
    \centering
    \caption{Standard ActionOPD and ablation training configurations. Each configuration corresponds to one training run. The number of samples refers to responses at one state, not independent complete trajectories.}
    \label{tab:training-summary}
    \small
    \setlength{\tabcolsep}{4pt}
    \renewcommand{\arraystretch}{1.15}
    \begin{tabularx}{\linewidth}{@{}>{\raggedright\arraybackslash}p{0.29\linewidth}>{\raggedright\arraybackslash}p{0.35\linewidth}>{\raggedright\arraybackslash}X@{}}
        \toprule
        \textbf{Setting} & \textbf{Transfer experiment} & \textbf{Tau ablations}\\
        \midrule
        Student & Qwen3-4B or Qwen3-8B & Qwen3-4B\\
        Action teacher & Qwen3.7-Flash or DeepSeek-V4-Flash & Qwen3-32B\\
        Training tasks & AWM/EnvScaler static-feasibility pools & Official train splits\\
        Collection steps reported & 100 & 50\\
        Task starts per step & 64: 59 AWM + 5 EnvScaler & 24: 8 per Tau domain\\
        Student samples $N$ & 4 & 4\\
        Teacher samples $K$ & 3 & 3\\
        Frequency bonus scale $\lambda$ & 0.5 & 0.5 for ActionOPD/random commit; 0 for binary reward\\
        Maximum agent decisions & 20 AWM; 40 EnvScaler & 20\\
        Optimizer & AdamW & AdamW\\
        Learning rate / schedule & $10^{-6}$ / constant, no warmup & $10^{-6}$ / constant, no warmup\\
        PPO epochs / minibatch size & 1 / 32 & 1 / 32\\
        PPO clip / dual clip & 0.2 / 3 & 0.2 / 3\\
        Advantage normalization & Group centering, batch whitening & Group centering, batch whitening\\
        Loss aggregation & Active-token mean & Active-token mean\\
        KL penalty / entropy bonus & Neither & Neither\\
        Outcome reward in loss & No & No\\
        \bottomrule
    \end{tabularx}
\end{table}

One collection step can involve multiple optimizer updates: one PPO epoch traverses the collected candidates in minibatches rather than updating once per task batch. Tau's 20-decision training limit is distinct from the native environment's 200-internal-transition budget. KD, OPD, and Self-ActionOPD share ActionOPD's optimizer, learning-rate, update, and sampling settings, but retain their respective objectives.

Main-experiment matcher and EnvScaler user-simulator settings are shared across all four student--teacher configurations. Tau training instead uses Qwen3.8-27B~\citep{qwen2026qwen38} for matching and user simulation; Tau evaluation uses Qwen3.5-9B~\citep{qwen2026qwen35}. Self-ActionOPD's private user summaries were prepared offline with DeepSeek-V4-Flash and reviewed, so auxiliary models remain part of its training setup.

In Table~\ref{tab:model-generation}, $T$, $p$, and $k$ denote temperature, top-$p$, and top-$k$, respectively. Response limits are in tokens, with 1k denoting 1,024 tokens.

\begin{table}[t]
    \centering
    \caption{Model roles and generation settings.}
    \label{tab:model-generation}
    \small
    \setlength{\tabcolsep}{3pt}
    \renewcommand{\arraystretch}{1.12}
    \begin{tabularx}{\linewidth}{@{}>{\raggedright\arraybackslash}p{0.13\linewidth}>{\raggedright\arraybackslash}X>{\raggedright\arraybackslash}p{0.16\linewidth}>{\raggedright\arraybackslash}p{0.19\linewidth}>{\centering\arraybackslash}p{0.12\linewidth}@{}}
        \toprule
        \multicolumn{5}{@{}l}{\textbf{Generation Settings}}\\
        \midrule
        \textbf{Experiment} & \textbf{Role and Model} & \textbf{Thinking} & \textbf{Sampling} & \textbf{Maximum Response}\\
        \midrule
        \multirow[t]{5}{=}{AWM/\newline EnvScaler\newline Training}
        & Student: Qwen3-4B/8B & Enabled & $T=0.6$; $p=0.95$; $k=20$ & 4k\\
        & Teacher: DeepSeek-V4-Flash & Enabled; effort max & --- & 4k\\
        & Teacher: Qwen3.7-Flash & Enabled & $T=0.6$; $p=0.95$ & 8k\\
        & Matcher: DeepSeek-V4-Flash & Disabled & $T=0$ & 128\\
        & User: DeepSeek-V4-Flash & Disabled & $T=1$ & ---\\
        \midrule
        \multirow[t]{5}{=}{Tau/Self-\newline ActionOPD\newline Training}
        & Student: Qwen3-4B & Enabled & $T=0.6$; $p=0.95$; $k=20$ & 4k\\
        & Teacher: Qwen3-32B & Enabled & $T=0.6$; $p=0.95$; $k=20$ & 8k\\
        & Self Teacher: Qwen3-4B & Enabled & $T=0.6$; $p=0.95$; $k=20$ & 4k\\
        & Matcher: Qwen3.8-27B & Disabled & $T=0.7$; $p=0.8$; $k=20$ & 32k\\
        & User: Qwen3.8-27B & Disabled & $T=0.7$; $p=0.8$; $k=20$ & 8k\\
        \midrule
        \multirow[t]{4}{=}{Evaluation}
        & Student: Qwen3-4B/8B & Enabled & Greedy ($T=0$) & 4k\\
        & Teacher: Qwen3.7-Flash & Enabled & $T=0.6$; $p=0.95$ & 32k\\
        & Teacher: DeepSeek-V4-Flash & Enabled; effort max & --- & 32k\\
        & Tau User: Qwen3.5-9B & Enabled & $T=1$ & 8k\\
        \bottomrule
    \end{tabularx}
\end{table}

\subsection{Evaluation protocol and score definitions}
\label{app:evaluation-protocol}

\paragraph{BFCL and ToolSandbox.}
Students and directly evaluated teachers use the same task subsets and interaction limits. BFCL includes 800 multi-turn tasks, 200 per category, with one trial and the native per-turn step limit of 20. ToolSandbox uses 490 strictly local scenarios, four scheduled trials each, and at most 30 messages. Similarity is averaged over all 1,960 scheduled trials.

\paragraph{AppWorld.}
AppWorld evaluation covers 168 Normal and 417 Challenge tasks for students and directly evaluated teachers, with one trial, 50 steps, and a 100,000-output-token budget per task. TGC is the fraction passing all task-goal tests; SGC measures completion of all variants within a scenario. Normal and Challenge scores are reported separately. The average uses TGC only. Shared task and interaction budgets do not imply matched generation budgets (Table~\ref{tab:model-generation}).

\paragraph{Tau transfer evaluation.}
Transfer evaluation covers all 50 Airline, 114 Retail, and 114 Telecom tasks in the official base splits, including the subsets used for Tau train/test comparisons. Transfer students were trained on AWM/EnvScaler, not these tasks. Evaluation uses native function calling and the generation settings in Table~\ref{tab:model-generation}. All models use four trials per task. For the six student models in Table~\ref{tab:main-results}, scores use trials 0--3 from the historical evaluations, without outcome-based trial selection.

All Tau evaluations in this paper disable natural-language assertions, including directly evaluated teachers, method comparisons, ablations, Self-ActionOPD, and the Privileged Teacher reference. Tau's evaluator supplies success labels under this scoring configuration, not the semantic matcher. Success therefore does not certify every utterance.

\paragraph{Tau method comparison, ablation, and Self-ActionOPD evaluation.}
Each model has four trials per task. Repeated Base and ActionOPD entries reuse the same results, not independent runs. Base scores partition the same trials 0--3 by official split task IDs. Airline, Retail, and Telecom train/test splits contain 30/20, 74/40, and 74/40 tasks. Training-set evaluation measures in-domain fit, not held-out generalization. Its episodes are distinct from training rollouts. Ordinary student evaluation excludes reference actions and private task criteria. The Privileged Teacher instead receives additional information and is not a same-input reference.

\Needspace{11\baselineskip}
\paragraph{Success and reliability.}
\label{app:score-definitions}

Following Tau's repeated-success metric~\citep{yao2025taubench}, for task $j$ with $n$ scheduled trials and $c_j$ successes, we compute
\begin{equation}
    \widehat p_j^{\,1}=\frac{c_j}{n},
    \qquad
    \widehat p_j^{\,4}=\frac{\binom{c_j}{4}}{\binom{n}{4}},
    \label{eq:tau-score-estimators}
\end{equation}
where $\binom{c_j}{4}=0$ when $c_j<4$. Each domain score averages these values over its tasks. With $n=4$, domain $\mathrm{pass}^{4}$ is the fraction of tasks successful in all four trials.

For the main table, let $B$ be BFCL accuracy, $S$ ToolSandbox similarity, $G_{\mathrm N}$ and $G_{\mathrm C}$ AppWorld Normal and Challenge TGC, and $P_d$ Tau $\mathrm{pass}^{1}$ in domain $d$. On a common percentage scale,
\begin{equation}
\begin{aligned}
    G &= \tfrac12(G_{\mathrm N}+G_{\mathrm C}),\\
    P &= \tfrac13(P_{\mathrm{Airline}}+P_{\mathrm{Retail}}+P_{\mathrm{Telecom}}),\\
    \mathrm{Avg} &= \tfrac14(B+S+G+P).
\end{aligned}
\label{eq:benchmark-macro}
\end{equation}
This study-specific summary weights benchmark families equally, without counting AppWorld SGC or Tau reliability again. Scores use one-decimal half-up rounding. Average gains are computed before display rounding, so subtracting displayed values can differ by 0.1 percentage points. Tau-only tables instead average the three domains separately for each metric and split; train and test are never pooled. Each configuration was trained once; repeated evaluation trials measure interaction variability, not variation across independent training runs.

\subsection{Tau baselines, ablations, and collection budgets}
\label{app:ablation-definitions}
\label{app:distillation-baselines}

All configurations use the same 178-task Tau training set and Qwen3-4B initialization. KD, OPD, and standard ActionOPD use Qwen3-32B as action teacher. Each step starts eight tasks per domain, and Tables~\ref{tab:distillation-comparison} and~\ref{tab:ablations} report step 50. Their ActionOPD rows reuse one run.

\paragraph{ActionOPD.}
The reference configuration uses \Eqref{eq:calibrated-reward} with $K=3$, $\lambda=0.5$, and uniform tie-breaking among maximum-reward candidates.

\paragraph{Validity Reward.}
Training assigns $+1$ to valid actions and $-1$ to invalid output or transfer-protocol violations. Teacher sampling and semantic matching remain active for ActionOPD's commitment rule, not training rewards. All-valid groups are masked as equal-reward groups, so this variant changes the number and composition of trainable groups as well as removing semantic reward distinctions.

\paragraph{Binary Reward.}
With the same three teacher samples, $\lambda=0$ gives $+1$ for any match, zero for valid non-matches, and $-1$ for invalid actions. This reward also governs commitment, changing both learning signals and ties relative to ActionOPD.

\paragraph{Random Commit.}
Rewards follow ActionOPD, but commitment is uniform over all four candidates, including invalid and zero-reward actions. There is no success-based resampling. All candidates in retained groups contribute to training.

A transfer notification without the preceding transfer call is treated as invalid output with reward $-1$. Equal-reward groups produce no learning signal.

\paragraph{KD: successful teacher-trajectory imitation.}
For each task start, Qwen3-32B independently generated four complete trajectories; only successful ones entered SFT. Following output imitation~\citep{kim2016sequence}, loss covers all teacher-generated reasoning and action/message tokens, while user turns and tool observations provide context only. Thus outcome labels filter demonstrations, but do not supply a policy-gradient reward.

\paragraph{OPD: sampled-token reverse-KL.}
OPD collected four student trajectories per task start. The teacher scored student-sampled tokens at their corresponding prefixes for sampled-token reverse-KL supervision~\citep{fu2026revisiting}. It requires teacher log-probabilities and a compatible token space, but not full-vocabulary distributions.

\paragraph{Collection budgets.}
KD and OPD collect four trajectories per task start; ActionOPD samples four candidates per state and executes one, while all candidates in retained groups contribute to training. Trajectory lengths, teacher generation/scoring, matching costs, and KD filtering differ. Equal task starts and collection steps therefore do not equalize retained examples, tokens, teacher queries, or compute.

\subsection{Self-ActionOPD: teacher inputs and parameter updates}
\label{app:self-aopd}

\paragraph{Current-weight teacher, fixed update targets.}
Self-ActionOPD used Qwen3-4B in both roles, with eight task starts per Tau domain per step. We report results at step 50. Four student candidates and three teacher samples were independently generated from the same current parameters. Teacher samples and matching results remained fixed during optimization. Subsequent collection used updated shared parameters. Self-teacher caches were cleared at each collection step.

Table~\ref{tab:model-generation} specifies generation settings. During Self-ActionOPD training, both roles used a 32,768-token context window and a 4,096-token response limit. Student and teacher prompts were capped at 24,576 and 28,672 tokens, respectively, providing an additional 4,096 tokens for teacher-only inputs. Teacher outputs and private prompt tokens were excluded from the student loss. Only student-generated candidate tokens received the action-level signal.

\paragraph{Teacher guidance and privileged information.}
The student received the domain policy, available tools, and observed interaction history. Additional teacher guidance covered immediate actions, preconditions, authorization, clarification, and avoiding repetition. Telecom device operations were rendered as user-facing instructions rather than user-tool syntax. This conversion added no private facts.

Privileged task information comprised reviewed summaries of private goals, constraints, preferences, and known/unknown information, including facts not yet disclosed in dialogue. It also supplied reference tool names and arguments, customer-side operations, target states, and required messages from evaluation criteria. These inputs did not replace required public confirmation or execution evidence. They were withheld from the student and matcher, although they could affect which student actions were rewarded. The trained student was evaluated without privileged inputs.

\end{document}